\RequirePackage{amsmath}
\documentclass[runningheads]{llncs}
\usepackage[T1]{fontenc}
\usepackage{comment}
\usepackage{multirow}
\usepackage[hyphens,spaces,obeyspaces]{url}
\usepackage[colorlinks,allcolors=blue]{hyperref}
\usepackage{float}
\usepackage{amsmath}%
\usepackage{graphicx}
\usepackage{tabularx}
\usepackage{amsfonts}
\usepackage{tabularx}
\begin{document}

\title{AACLiteNet: A Lightweight Model for Detection of Fine-Grained Abdominal Aortic Calcification}
\titlerunning{AACLiteNet for AAC detection}
\author{Zaid Ilyas\inst{1,2} \and Afsah Saleem\inst{1,2} \and David  Suter\inst{1,2} \and Siobhan Reid\inst{6} \and John T. Schousboe \inst{4} \and William D. Leslie \inst{5}  \and  Joshua R. Lewis \inst{1,2} \and Syed Zulqarnain Gilani \inst{1,2,3}}
\authorrunning{Z. Ilyas et al.}

\institute{Centre for AI \& ML, School of Science, Edith Cowan University, Australia \and Nutrition and Health Innovation Research Institute, Edith Cowan University,
Australia \and 
Computer Science and Software Engineering, The University of Western Australia  \and  
Park Nicollet Clinic and HealthPartners Institute, Minneapolis,USA \and 
Department of Medicine and Radiology, University of Manitoba, Canada \and
Department of Electrical and Computer Engineering, University of Manitoba, Winnipeg, Canada\\
\email{z.ilyas@ecu.edu.au}}

\maketitle      
\begin{abstract}
Cardiovascular Diseases (CVDs) are the leading cause of death worldwide, taking 17.9 million lives annually. Abdominal Aortic Calcification (AAC) is an established marker for CVD, which can be observed in lateral view Vertebral Fracture Assessment (VFA) scans, usually done for vertebral fracture detection. Early detection of AAC may help reduce the risk of developing clinical CVDs by encouraging preventive measures. Manual analysis of VFA scans for AAC measurement is time-consuming and requires trained human assessors. Recently, efforts have been made to automate the process; however, the proposed models are either low in accuracy, lack granular-level score prediction, or are too heavy in terms of inference time and memory footprint. Considering all these shortcomings of existing algorithms, we propose `AACLiteNet', a lightweight deep learning model that predicts both cumulative and granular level AAC scores with high accuracy, and also has a low memory footprint, and computation cost (Floating Point Operations (FLOPs)). The AACLiteNet achieves a significantly improved one-vs-rest average accuracy of 85.94$\%$ as compared to the previous best 81.98$\%$, with 19.88 times less computational cost and 2.26 times less memory footprint, making it implementable on portable computing devices.      

\end{abstract}

\section{Introduction}
\begin{figure}[t]
\centering
\includegraphics[scale=0.3]{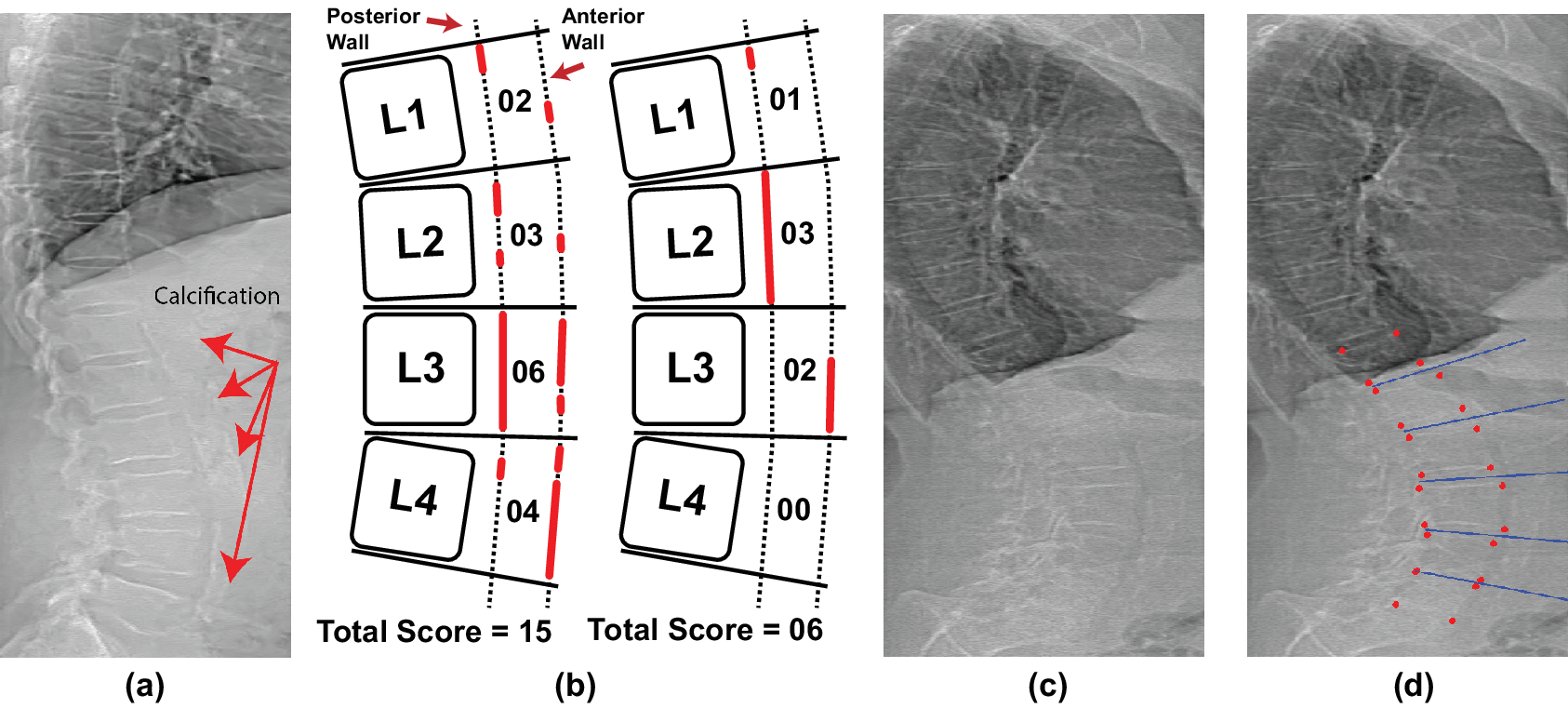}
\caption{(a) The red arrows point to calcified pixels. (b) Two examples of AAC scoring. (c) DXA image with Poorly demarcated vertebral boundaries. (d) Use of guides for AAC measurement using Kauppila AAC-24 Point Scoring Method \cite{r17}. } \label{fig4}
\end{figure}

Cardiovascular Diseases (CVDs) are the leading cause of death worldwide, affecting 17.9 million people each year \cite{r1}. These diseases are commonly associated with heart and blood vessels~\cite{r2}. Their identification and timely treatment can prevent premature deaths. Atherosclerosis, a condition that is a precursor to CVDs, results in the calcification of blood vessels. The Abdominal aorta is one of the first vascular beds where this condition is known to manifest \cite{r3,r4}. Abdominal Aortic Calcification (AAC) is a stable marker of atherosclerosis and can help predict future CVD events \cite{r5,r6,r7,r8}. Early detection of AAC may help in encouraging preventive measures to avoid adverse outcomes related to CVD including premature death. 
AAC can be detected using different imaging modalities like Computed Tomography (CT) \cite{r9}, Digital X-Ray Imaging \cite{r10}, Dual-Energy X-Ray Absorptiometry (DXA) \cite{r6,r7,r10,r11}, Magnetic Resonance Imaging (MRI) \cite{r12}, and Ultrasound \cite{r12}. DXA is the least expensive modality in terms of cost and radiation exposure to patients, making it the tool of choice. However, the detection of AAC from Vertebral Fracture Assessment (VFA) scans (captured using DXA) can be challenging as these images are of low resolution, can have artifacts, and poorly-demarcated vertebral boundaries (an example shown in  Fig. \ref{fig4}(c)). AAC scoring is mostly done using the AAC-24 point semi-quantitative scoring method developed by Kauppila LI et al. \cite{r17}. In this method, the abdominal aorta adjacent to L1 to L4 vertebrae is divided into eight sections i.e. L1-anterior to L4-anterior, and L1-posterior to L4-posterior. In each section, a score of one is given if AAC $\leq$ 1/3 of vertebral length, two if 1/3$>$ AAC $\leq$ 2/3 of vertebral length, and three if AAC$>$2/3 of vertebral length. This rule is applied to all eight sections, giving a maximum possible cumulative score of 24. Fig. \ref{fig4}(b) illustrates two examples of AAC calculation using the Kauppila-24 method from DXA images.\par
Manual assessment of VFA scans for AAC is time-consuming and expensive. Unfortunately, the automatic detection of AAC from VFA images has received limited attention. The initial work in this domain is from Elmasri et al. \cite{r14} and Chaplin et al. \cite{r13} measured a cumulative AAC score using statistical models. These methods used non-standardized AAC scores for training and didn't focus on the granular-level scores (AAC scores against L1 to L4 vertebrae).  Reid et al. \cite{r15} trained a deep learning model on 1,100 GE-Lunar Prodigy and iDXA VFA images using the Inception-ResNet2 backbone-based architecture. Their model predicted AAC as a regression score between 0-24: which was then classified into three risk categories, namely Low-Risk Class (AAC score: 0-1), Moderate-Risk Class (AAC score: 2-5), and High-Risk Class (AAC $\ge$ 6). The authors did not perform k-fold validation and reported an average one-vs-all test set accuracy of 88.1$\%$ and an $R^{2}$ coefficient of 0.86. Although the proposed model performed comparatively better than its predecessors, it was computationally (60.80 Giga Floating Point Operations (FLOPs)) and memory-wise heavy (57.46 Million parameters). Recently, Gilani et al. \cite{r16} proposed a model to predict granular level scores from VFA DXA images. They used 1,916 bone density images to train a Long Short Term Memory (LSTM) based model using transfer learning. Using 10-fold cross-validation, they reported an average one-vs-rest accuracy of 81.98\%. However, their model is also quite heavy (68.86 Million parameters) and computationally expensive (64.03 Giga FLOPs). We argue that an ideal model should excel in both aspects, i.e. exhibit strong performance on granular AAC scores, while also being fast and resource-efficient for suitability on mobile computing devices. 

We address the shortcomings of the methods mentioned above by proposing a novel lightweight model, `AACLiteNet', which detects granular-level AAC scores with state-of-the-art performance.
The novelty in our work is the design of a lightweight Convolutional Neural Network (CNN) model with an efficient global attention mechanism (inspired by the transformer architecture \cite{r18,r_deit}) that is trainable on a small-sized dataset and can predict both cumulative and granular AAC scores in a single head. AACLiteNet was successfully trained on 1,916 iDXA GE Machine images \cite{r16}. Our model achieved a state-of-the-art one vs. rest accuracy of 87.53$\%$ (Low category), 80.22$\%$ (Medium category), and 90.08$\%$ (High Category). Additionally, our model has 19.88 times less computational cost and 2.26 times less memory footprint as compared to the previous state-of-the-art model \cite{r16}. The code for our work can be found at \cite{c1}. Our contributions in this work can be summarized as follows:
\begin{itemize}
\item A lightweight CNN model with an efficient global attention mechanism, successfully trained on a small-sized dataset achieving state-of-the-art accuracy with significantly less memory footprint and computational cost. 
\item Framing the problem as a combination of regression and Multi-Class Multi-Label classification tasks, reinforcing each other during the training phase eliminating the need for separate decoder models. 
\item We showed that Hazard ratios for Major Acute Cardiovascular Event (MACE) outcome prediction by our proposed model overlapped with those from trained human assessors.
\end{itemize}

\section{Proposed Framework}\label{Proposed Framework}
\begin{figure}[t]

\centering

\includegraphics[width=\textwidth]{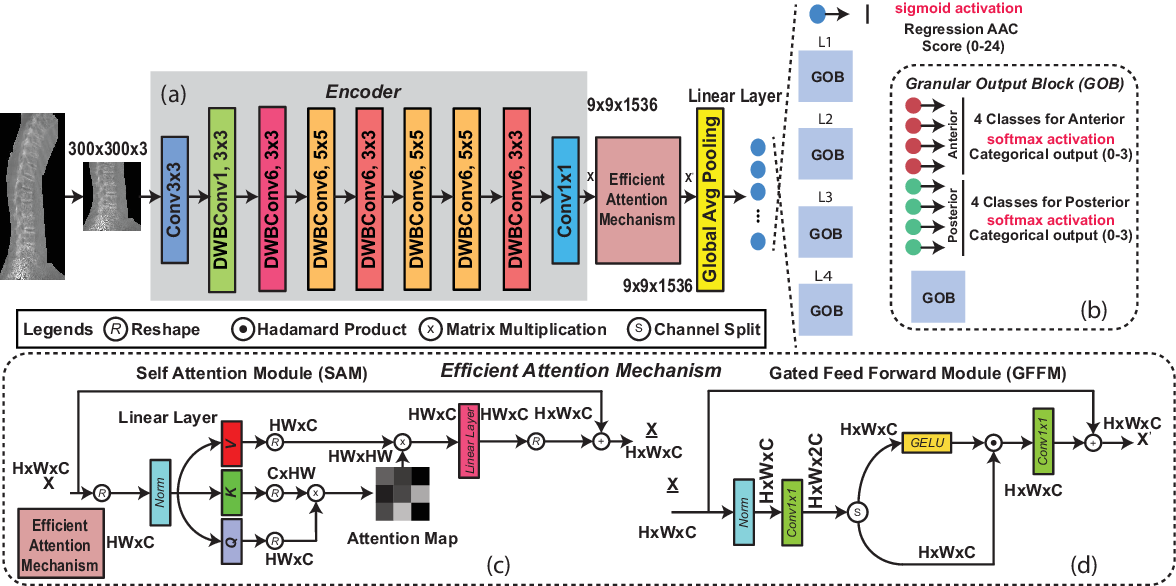}

\caption{(a) CNN Encoder with multiple 2D simple and depthwise convolution layers. (b) Granular Output Block with four outputs each for anterior and posterior sections of one vertebra. (c) SAM Block for Self-Attention. (d) GFFM Blocks for Channel Attention using Gating Mechanism.} \label{fig1}
\end{figure}

Our proposed framework, AACLiteNet, is shown in Fig. \ref{fig1}(a). Firstly, we describe the network architecture, before explaining the rationale for the design. AACLiteNet consists of a preprocessing unit, a CNN encoder module, an efficient attention mechanism block, and an output head that predicts cumulative and granular scores. The preprocessing module performs necessary preprocessing and crops the image to the size of $\mathbb{R}^{300 \times 300 \times 3}$. This cropped image is processed by the encoder module, which performs multiple 2D convolution operations with different kernel sizes at various depth levels of the network to learn complex feature representations. Inspired by the EfficietNet family of CNN models \cite{r20}, we designed our encoder by incorporating  Depthwise Convolution (DWC) operations in addition to conventional 2D convolution operations. DWC operation performs a separate 2D convolution operation on each channel of the feature map which greatly reduces the network's computation cost and memory footprint. We used DWC in the bottleneck blocks of our encoder network which are abbreviated as `DWBConv'. In these blocks, the input feature map is first expanded along the channels dimension, then DWC operations are performed, which process each channel separately independent of each other, and then the resultant feature map is squeezed along the channels back to the size of the original feature map. The generated feature map is then added element-wise to the input feature map. All the processes involved in the bottleneck block help the model in learning complex feature representations with fewer operations and parameters and also helps in better convergence during training. The generated feature maps from the encoder block are rich in complex spatial feature representation, however, to further add global context among these features, we use a self-attention mechanism (inspired by Transformer architecture \cite{r18,r_deit}) that calculates long-range dependencies among spatial features. Given a preprocessed input image of size $\mathbb{R}^{300 \times 300 \times 3}$, the encoder module outputs a feature map of size $\mathbb{R}^{9 \times 9 \times 1536}$ that is fed to the Self-Attention Module (SAM) in the `Efficient Attention Mechanism' block of our network. The SAM block flattens the feature map to the shape $\mathbb{R}^{81 \times 1536}$ and then calculates the self-attention among 81 features considering 1536 as the embedding length of each feature. This self-attention is important as it allows the model to selectively focus on different parts of an input sequence (of length 81 in the present case) by computing a weighted sum of the other parts based on their similarity or relevance to the current part.  The SAM module is incorporated into the framework keeping the best possible trade-off between computational and memory cost vs. performance of the network. This placement is important as a linear increase in the number of spatial features quadratically increases the computation requirements of the self-attention mechanism. The pipeline of the SAM block is shown in Fig. \ref{fig1}(c). It generates Queries (\textbf{Q}), Keys (\textbf{K}), and Values (\textbf{V}) embeddings of size $\mathbb{R}^{81 \times 1536}$ from the Linear Normalized input feature map \textbf{Y} and then performs following operations to calculate self-attention:
\begin{equation}
    \textbf{Q}, \textbf{K}, \textbf{V} = W_{Q}\textbf{Y}, W_{K}\textbf{Y}, W_{V}\textbf{Y},\text{ where } \mbox{\textbf{Y} = LayerNorm(\textbf{X})}
\end{equation}
\begin{equation}
    \textbf{\underbar{X}} = \mbox{Softmax}(\textbf{Q} \cdot \textbf{K}^{T}/\alpha) \cdot \textbf{V} + \textbf{X}, 
    \text{ \textit{where} Softmax}(x_{i}) = \frac{\exp(x_i)}{\sum_j \exp(x_j)}
\end{equation}

where $W_{Q}$, $W_{K}$, and $W_{V}$ are the learnable parameters to transform the input feature map $\textbf{Y}$ into \textbf{Q}, \textbf{K}, and \textbf{V} embeddings, and $\alpha$ is a learnable parameter to control the intensity of self-attention. The SAM calculates self-attention amongst 81 spatial features. However, to add attention to 1536 channels as well, our network's pipeline passes the output from SAM through a novel Gated Feed-Forward Module (GFFM). Unlike conventional Feed-Forward Networks (FFN), used in Transformer architectures \cite{r18,r_deit}), this module first expands the feature map along the channel dimension and then splits it into two parts of size $\mathbb{R}^{9 \times 9 \times 1536}$ each. One part is passed through a Gaussian Error Linear Unit (GELU) activation layer and then multiplied element-wise (Hadamard Product) to the other part. This type of attention mechanism serves as a gating technique to control the flow of information. More weight is added to relevant/ critical information while irrelevant information is suppressed. The GFFM can be mathematically formulated as:    
\begin{equation}
    \textbf{X}^{'} = W_{o}(\phi(W_{1}\textbf{\underbar{Z}})\odot W_{2}\textbf{\underbar{Z}}) + \textbf{\underbar{X}}, \text{ where \mbox{\textbf{Z} = LayerNorm(\textbf{\underbar{X}})}}
\end{equation}

where $W_{1}$ and $W_{2}$ are the learnable parameters used to split the input feature map $\textbf{\underbar{Z}}$ along channels, $\phi$ is the GELU activation function, $W_{0}$ is the output projection matrix, and $\odot$ is the element-wise multiplication. Next, the Global Average Pooling (GAP) operations are performed on the output of GFFM. GAP output is flattened and fed to a Linear Layer, which outputs 33 values. The first output branch of the Linear Layer is fed to a Sigmoid activation function which predicts the regression score, and the outputs of the remaining 32 branches are tackled in a Multi-Class Multi-Label manner i.e. 32 values are divided into eight sections, each of size four representing outputs of anterior and posterior sections of four vertebrae L1, L2, L3, and L4. The four values of all eight sections are fed to Softmax activation layers separately. Each of the eight sections predicts one of four classes representing AAC scores in the range of 0-3.  

\section{Experiments and Results}\label{Experiments and Results}
We used 1,916 VFA images from an iDXA GE machine \cite{r16} with a resolution of 1600 $\times$ 300 pixels. The dataset consists of 829 low-risk, 445 medium-risk, and 642 high-risk images with AAC score distribution for all the vertebrae L1-L4 highly skewed towards the zero scores. The input image is cropped 50$\%$ from the top, 40$\%$ from the left, and 10$\%$ from the right. Unlike \cite{r15,r16}, we used smaller-sized images (300$\times$300 pixels) in our framework to reduce computational cost and memory footprint. The image values are normalized between 0-1. Data augmentation techniques like scaling, translation, rotation and shear were used in our experiments. 10-fold Stratified Cross Validation was used, with each fold having 1,724 train images and 192 test images. The network training was done using both regression and granular scores.  The weighted Mean Square Error (MSE) loss function was used for the regression output and for granular outputs, eight weighted Categorical Cross Entropy (CCE) loss functions for vertebrae L1-L4 in a Multi-Class Multi-Label classification manner. The loss function $L_{Total}$ used is formulated as:
\begin{equation}
    L_{Total} = \left ( w_{Reg} \cdot L_{Reg} + \sum_{i=1}^{4}w_{Ai}\cdot L_{Ai} + \sum_{i=1}^{4}w_{Pi}\cdot L_{Pi} \right )/2
\end{equation}
where $L_{Reg}$ is the MSE Loss for cumulative AAC score, $w_{Reg}$ is the weighing factor to balance class distribution, $L_{Ai}$ and $L_{Pi}$ are the CCE Losses for Anterior and Posterior levels for all four vertebrae ranging L1 to L4, with respective weights $w_{Ai}$ and $w_{Pi}$ for class imbalance correction. We used a batch size of 20, a learning rate of 5e$^{-4}$, and Adam Optimizer \cite{kingma} for training.

\begin{figure}[h]
\centering
\includegraphics[width=\textwidth]{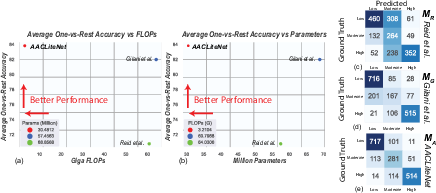}
\caption{(a) Average one-vs-rest accuracy vs FLOPs. (b) Average one-vs-rest accuracy vs parameter count. (c) Confusion Matrix Reid et al. \cite{r15} (d) Confusion Matrix Gilani et al. \cite{r16} (e) Confusion Matrix AACLiteNet.} \label{fig_cm}
\end{figure}

\begin{table}[t]

\caption{Performance Comparison in one-vs-rest configuration. NPV / PPV stands for Negative/ Positive Predicted Value. $M_{R}$ represents work of Reid et al. \cite{r15}, $M_{G}$ represents work of Gilani et al. \cite{r16}, and $M_{A}$ represents our proposed model AACLiteNet.  }\label{tab2}
\resizebox{\columnwidth}{!}{
\begin{tabular}{|cccccccccc|ccc|}

\hline
\multicolumn{1}{|c|}{\multirow{2}{*}{}} & \multicolumn{3}{c|}{Low Category}                                                             & \multicolumn{3}{c|}{Medium Category}                                                                   & \multicolumn{3}{c|}{High Category}                                                & \multicolumn{3}{c|}{Mean}                                                \\ \cline{2-13} 
\multicolumn{1}{|c|}{}                  & \multicolumn{1}{c|}{$M_{R}$}    & \multicolumn{1}{c|}{$M_{G}$}    & \multicolumn{1}{c|}{$M_{A}$}             & \multicolumn{1}{c|}{$M_{R}$}    & \multicolumn{1}{c|}{$M_{G}$}             & \multicolumn{1}{c|}{$M_{A}$}             & \multicolumn{1}{c|}{$M_{R}$}    & \multicolumn{1}{c|}{$M_{G}$}             & $M_{A}$             & \multicolumn{1}{c|}{$M_{R}$}    & \multicolumn{1}{c|}{$M_{G}$}    & $M_{A}$             \\ \hline
\multicolumn{1}{|c|}{Accuracy}         & \multicolumn{1}{c|}{71.14} & \multicolumn{1}{c|}{82.52} & \multicolumn{1}{c|}{\textbf{87.53}} & \multicolumn{1}{c|}{62.06} & \multicolumn{1}{c|}{75.52}          & \multicolumn{1}{c|}{\textbf{80.22}} & \multicolumn{1}{c|}{79.12} & \multicolumn{1}{c|}{87.89}          & \textbf{90.08} & \multicolumn{1}{c|}{70.77} & \multicolumn{1}{c|}{81.98} & \textbf{85.94} \\
\multicolumn{1}{|c|}{Sensitivity}       & \multicolumn{1}{c|}{55.49} & \multicolumn{1}{c|}{86.37} & \multicolumn{1}{c|}{\textbf{86.54}} & \multicolumn{1}{c|}{59.33} & \multicolumn{1}{c|}{37.53}          & \multicolumn{1}{c|}{\textbf{63.15}} & \multicolumn{1}{c|}{54.83} & \multicolumn{1}{c|}{\textbf{80.22}} & 80.09          & \multicolumn{1}{c|}{56.55} & \multicolumn{1}{c|}{68.04} & \textbf{76.59} \\
\multicolumn{1}{|c|}{Specificity}       & \multicolumn{1}{c|}{83.07} & \multicolumn{1}{c|}{79.58} & \multicolumn{1}{c|}{\textbf{88.32}} & \multicolumn{1}{c|}{62.88} & \multicolumn{1}{c|}{\textbf{87.02}} & \multicolumn{1}{c|}{85.42}          & \multicolumn{1}{c|}{91.37} & \multicolumn{1}{c|}{91.76}          & \textbf{95.15} & \multicolumn{1}{c|}{79.11} & \multicolumn{1}{c|}{86.12} & \textbf{89.63} \\
\multicolumn{1}{|c|}{NPV}               & \multicolumn{1}{c|}{70.99} & \multicolumn{1}{c|}{88.45} & \multicolumn{1}{c|}{\textbf{89.56}} & \multicolumn{1}{c|}{83.63} & \multicolumn{1}{c|}{82.16}          & \multicolumn{1}{c|}{\textbf{88.48}} & \multicolumn{1}{c|}{80.06} & \multicolumn{1}{c|}{90.20}          & \textbf{90.47} & \multicolumn{1}{c|}{78.23} & \multicolumn{1}{c|}{86.93} & \textbf{89.51} \\
\multicolumn{1}{|c|}{PPV}               & \multicolumn{1}{c|}{71.43} & \multicolumn{1}{c|}{76.33} & \multicolumn{1}{c|}{\textbf{85.01}} & \multicolumn{1}{c|}{32.59} & \multicolumn{1}{c|}{46.65}          & \multicolumn{1}{c|}{\textbf{56.65}} & \multicolumn{1}{c|}{76.19} & \multicolumn{1}{c|}{83.06}          & \textbf{89.25} & \multicolumn{1}{c|}{60.07} & \multicolumn{1}{c|}{68.68} & \textbf{76.97} \\ \hline
\multicolumn{10}{|c|}{Pearson Correlation ($r$)}                                                                                                                                                                                                                                                                                       & \multicolumn{1}{c|}{0.65}  & \multicolumn{1}{c|}{0.84}  & \textbf{0.89}  \\
\multicolumn{10}{|c|}{Coefficient of Determination ($R^{2}$)}                                                                                                                                                                                                                                                                                  & \multicolumn{1}{c|}{0.58}  & \multicolumn{1}{c|}{0.69}  & \textbf{0.79}  \\
\multicolumn{10}{|c|}{Computational Cost / Floating Point Operations (FLOPs) (Giga)}                                                                                                                                                                                                                                                                      & \multicolumn{1}{c|}{60.80} & \multicolumn{1}{c|}{64.03} & \textbf{3.22}  \\
\multicolumn{10}{|c|}{Memory Footprint / Parameter Count (Million)}                                                                                                                                                                                                                                                                                     & \multicolumn{1}{c|}{57.46} & \multicolumn{1}{c|}{68.86} & \textbf{30.49} \\ \hline
\end{tabular}}
\end{table}
\begin{table}[]
\centering
\caption{Granular scores accuracy ($\%$) (One vs. Rest) (Left). Ablation Study for using different output labels for training i.e. Regression only, granular only, and combined (Right).}\label{tab3}
\resizebox{\columnwidth}{!}{\begin{tabular}{|ccccc||cccccccccc|}
\hline
\multicolumn{5}{|c||}{Granular Scores Accuracy}                                                                                                                & \multicolumn{10}{c|}{Possible configurations of labels for Training}                                                                                                                                                                                   \\ \hline
\multicolumn{1}{|c|}{\multirow{2}{*}{Vertebra}} & \multicolumn{2}{c|}{Anterior}                                       & \multicolumn{2}{c||}{Posterior}                 & \multicolumn{1}{c|}{\multirow{2}{*}{Labels}} & \multicolumn{3}{c|}{\multirow{2}{*}{Accuracy ($\%$)}}                                    & \multicolumn{3}{c|}{\multirow{2}{*}{Sensitivity ($\%$)}}                                 & \multicolumn{3}{c|}{\multirow{2}{*}{Specificity ($\%$)}}            \\ \cline{2-5}
\multicolumn{1}{|c|}{}                          & \multicolumn{1}{c|}{$M_{G}$}     & \multicolumn{1}{c|}{$M_{A}$}     & \multicolumn{1}{c|}{$M_{G}$}     & $M_{A}$     & \multicolumn{1}{c|}{}                        & \multicolumn{3}{c|}{}                                                             & \multicolumn{3}{c|}{}                                                             & \multicolumn{3}{c|}{}                                        \\ \hline
\multicolumn{1}{|c|}{L1}                        & \multicolumn{1}{c|}{71}          & \multicolumn{1}{c|}{\textbf{73}} & \multicolumn{1}{c|}{\textbf{72}} & \textbf{72} & \multicolumn{1}{c|}{}                        & \multicolumn{1}{c|}{Low}  & \multicolumn{1}{c|}{Med}  & \multicolumn{1}{c|}{High} & \multicolumn{1}{c|}{Low}  & \multicolumn{1}{c|}{Med}  & \multicolumn{1}{c|}{High} & \multicolumn{1}{c|}{Low}  & \multicolumn{1}{c|}{Med}  & High \\ \cline{7-15} 
\multicolumn{1}{|c|}{L2}                        & \multicolumn{1}{c|}{\textbf{73}} & \multicolumn{1}{c|}{\textbf{73}} & \multicolumn{1}{c|}{65}          & \textbf{69} & \multicolumn{1}{c|}{Regression}                & \multicolumn{1}{c|}{86.0} & \multicolumn{1}{c|}{77.0} & \multicolumn{1}{c|}{89.0} & \multicolumn{1}{c|}{86.0} & \multicolumn{1}{c|}{57.0} & \multicolumn{1}{c|}{76.0} & \multicolumn{1}{c|}{86.0} & \multicolumn{1}{c|}{83.0} & 95.0 \\
\multicolumn{1}{|c|}{L3}                        & \multicolumn{1}{c|}{70}          & \multicolumn{1}{c|}{\textbf{75}} & \multicolumn{1}{c|}{69}          & \textbf{74} & \multicolumn{1}{c|}{Granular}                & \multicolumn{1}{c|}{83.3} & \multicolumn{1}{c|}{78.9} & \multicolumn{1}{c|}{87.8} & \multicolumn{1}{c|}{90.7} & \multicolumn{1}{c|}{33.0} & \multicolumn{1}{c|}{83.7} & \multicolumn{1}{c|}{92.8} & \multicolumn{1}{c|}{77.6} & 89.8 \\
\multicolumn{1}{|c|}{L4}                        & \multicolumn{1}{c|}{72}          & \multicolumn{1}{c|}{\textbf{76}} & \multicolumn{1}{c|}{70}          & \textbf{72} & \multicolumn{1}{c|}{Combined}                & \multicolumn{1}{c|}{87.5} & \multicolumn{1}{c|}{80.2} & \multicolumn{1}{c|}{90.1} & \multicolumn{1}{c|}{86.5} & \multicolumn{1}{c|}{63.1} & \multicolumn{1}{c|}{80.2} & \multicolumn{1}{c|}{88.3} & \multicolumn{1}{c|}{87.0} & 95.1 \\ \hline
\end{tabular}}
\end{table}

\begin{figure}[htb]
\centering
\includegraphics[scale=0.5]{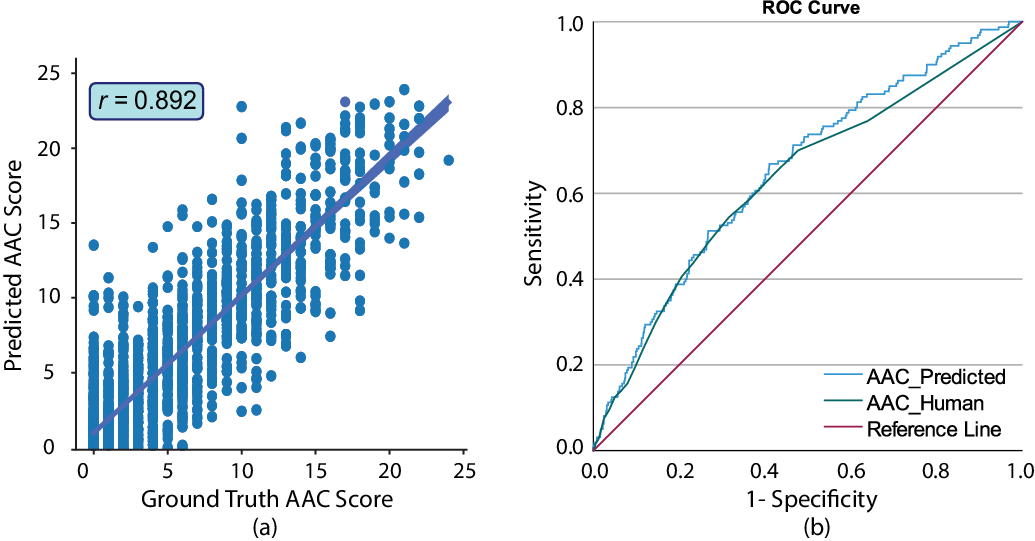}
\caption{(a) Pearson Correlation of AACLiteNet AAC score with human ground truth. (b) ROC Curves for MACE prediction.} \label{fig_pearson}
\end{figure}

\noindent{\textbf{Results}:} The comparative analysis of AACLiteNet with Reid et al. \cite{r15} and Gilani et al. \cite{r16} is shown in Table \ref{tab2}, where $M_{R}$ represents the results of Reid et al. \cite{r15}, $M_{G}$ depicts Gilani et al. \cite{r16}, and $M_{A}$ represents our AACLiteNet. AACLiteNet achieves a significantly improved one-vs-all average accuracy of 85.94$\%$, compared to the previous best of 81.98$\%$\cite{r16}. The average 3-class classification accuracy of our model is 78.91$\%$ which is significantly better than 72.80$\%$ of \cite{r16} and 55.8$\%$ of~\cite{r15}. Fig. \ref{fig_cm}(a-b) shows AACLiteNet has a significant reduction in memory footprint, i.e. 46.94$\%$  compared to \cite{r15}, and 55.72$\%$ compared to \cite{r16}, and a computation cost reduction (FLOPs) of 94.97$\%$ over \cite{r16}. The confusion matrices of the three models are shown in Fig. \ref{fig_cm}(c-e). Our model has a lower misclassification rate, especially from high to medium and medium to low categories. Our model also improves the one-vs-rest average granular score accuracy for L1 to L4 anterior and posterior sections, as shown in Table \ref{tab3} (left). Finally, we show (Fig. \ref{fig_pearson}(a)) that AACLiteNet has a better correlation with human-annotated granular AAC scores. We report Pearson Correlation of $0.89$ compared to $0.84$ of \cite{r16} ($p<<0.001$).\par

\noindent\textbf{Ablation Studies}: We used three possible configurations of labels to train our model. Table~\ref{tab3} (right) shows that the best results were produced by the third configuration (i.e. predicting joint AAC-24 and granular scores).\par

\noindent\textbf{Clinical Significance}: Among a cohort of 1,877 people, 160 suffered a Major Acute Cardiovascular Events (MACE) outcome. The distribution of the cohort is such that 827 (44.1$\%$) scans belonged to the low AAC group, 488 (26.0$\%$) to the moderate AAC group, and 562 (29.9$\%$) to the high AAC group. MACE events occurred in 4.7$\%$ of the low AAC group, 8.0$\%$ in the moderate AAC group, and 14.6$\%$ in the high AAC group. In both the moderate and high AAC groups, the hazard ratios for MACE prediction (versus low-risk AAC) by our model overlapped with hazards ratios for AAC scores from humans i.e. moderate (model calculated) 1.54 95$\%$CI 0.98-2.40 vs (human calculated) 1.15 95$\%$CI 0.72-1.84; high (model calculated) 2.77 95$\%$ CI 1.87-4.12 vs. 2.32 95$\%$ CI 1.59-3.38 (human calculated). Fig. \ref{fig_pearson}(b) compares the Area Under the Curve (AUC) of our model with human-annotated AAC scores for MACE prediction. After adjusting for the correlation using the McNeil-Hanley method \cite{r_curves} AACLiteNet led to an incremental improvement in the AUC compared to human AAC-24 (p < 0.001). (0.66 95$\%$CI 0.61-0.70 vs 0.63 95$\%$CI 0.59-0.68). 

\section{Conclusion}\label{Conclusion}
We propose AACLiteNet, a lightweight CNN model with an efficient global attention mechanism that predicts cumulative and granular AAC scores from VFA images with high accuracy. Its performance surpassed its predecessors with a significant reduction in the memory footprint and computation cost. Moreover, the hazard ratios calculated for MACE prediction by AACLiteNet compare favorably with ratios calculated by trained human assessors. A limitation of our work is that we could only test AACLiteNet on granular-level annotated scans of iDXA GE machine scans. No other annotated data was available from any other source. In the future, we plan to test the effectiveness of our proposed model on Hologic machine scans. 

\begin{thebibliography}{30}
\bibitem{c1}
To be released after acceptance of the paper.
\end{thebibliography}
\end{document}